\documentclass[11pt,a4paper]{article}
\usepackage[hyperref]{acl2021}
\usepackage{times}
\usepackage{latexsym}

\usepackage{graphicx}

\usepackage{CJKutf8}
\usepackage{tabularx}
\usepackage{chngcntr}

\usepackage[utf8]{inputenc}
\usepackage[T1]{fontenc} %
\usepackage{multirow}
\usepackage{booktabs}
\usepackage{amsmath}
\usepackage{amsfonts}

\usepackage{seqsplit}
\usepackage{newunicodechar}
\usepackage{microtype}
\usepackage{cleveref}

\crefformat{section}{\S#2#1#3}
\crefformat{subsection}{\S#2#1#3}
\crefformat{subsubsection}{\S#2#1#3}
\aclfinalcopy 
\def\aclpaperid{2} 

\title{ Time-Aware Ancient Chinese Text Translation and Inference  }
\author{ 
  $\star$Ernie Chang\textsuperscript{{\begin{CJK}{UTF8}{gbsn}文\end{CJK}}}, $\circ$Yow-Ting Shiue\textsuperscript{{\begin{CJK}{UTF8}{gbsn}文\end{CJK}}}, $\star$Hui-Syuan Yeh, $\star$Vera Demberg \\
  $\star$Dept. of Language Science and Technology, Saarland University \\
  $\circ$Dept. of Computer Science, University of Maryland, College Park \\
    {\tt \{cychang\}@coli.uni-saarland.de, ytshiue@cs.umd.edu}
  \\ 
}

\date{}

\begin{document}
\maketitle
\begin{CJK}{UTF8}{gbsn}

\renewcommand*{\thefootnote}{\roman{footnote}}
\footnotemark[0]
\footnotetext[0]{\textsuperscript{{\begin{CJK}{UTF8}{gbsn}文\end{CJK}}}These authors contributed equally.}
\renewcommand*{\thefootnote}{\arabic{footnote}}
\setcounter{footnote}{0}

\begin{abstract}

In this paper, we aim to address the challenges surrounding the translation of ancient Chinese text:
(1) The \emph{linguistic gap} due to the difference in eras results in translations that are poor in quality, and
(2) most translations are missing the contextual information that is often very crucial to understanding the text. 
To this end, we improve upon past translation techniques by proposing the following:
We reframe the task as a \emph{multi-label prediction task} where the model predicts both the translation and its particular era. 
We observe that this helps to bridge the linguistic gap as chronological context is also used as auxiliary information.
We validate our framework on a parallel corpus annotated with chronology information and show experimentally its efficacy in producing quality translation outputs.
We release both the code and the data\footnote{\url{https://github.com/orina1123/time-aware-ancient-text-translation}} for future research.
\end{abstract}

\section{Introduction}

The Chinese language inherits a lot of phrases from ancient time \cite{bao2008illustrations,liu2019formulaic} and is spoken by roughly $1.3$ billion native speakers.
However, the language's ancient variant (or \emph{ancient Chinese}) is mastered by a few and proved to be a bottleneck in understanding the essence of the Chinese culture. 
Building a translation system from the ancient Chinese to the modern text thus serves a few important purposes:
(I) The ancient Chinese is considered as an essential part of the curriculum in all of the Chinese-speaking regions\footnote{This includes the mainland China, Taiwan, Singapore, Malaysia, etc. }, so an ancient Chinese translation system can be used to bolster the immediate understanding of ancient texts.
(II) Further, the translation system can help to settle the linguistic debate with regard to the era of origin of an independent segment of text.
This is especially useful for the identification of a  discovered artifacts where carbon dating cannot pinpoint the exact era, but where their linguistic features can formulate a clear-cut dynasty or time period.

However, it is not without challenge in constructing such translation systems. 
One primary obstacle lies in the extensive timeline where ancient texts can be derived -- one segment of ancient text can come from the Pre-Qin (先秦) era, and another coming from the Song dynasty (宋朝), which are roughly about $700$ years apart.
This gap witnessed a drastic evolution of linguistic properties where the usage of phrases became imbued with different meanings.
Besides, different eras often consist of various amounts of available data, and thus the same translation model training will be exposed to data imbalance, which complicates the design of the translation systems and limits their generalizability.
On the other hand, past attempts at building such translation systems yield poor performance that renders them practically unusable as-is in the practical settings~\cite{zhang2018automatic,liu2019ancient} -- these efforts are still largely limited as parallel data is scarce for some eras.

Recent advances in machine translation and text style transfer/generation utilize semi-supervised techniques to tackle similar challenges by aligning latent representations from different styles for the low resource scenarios~\cite{shen2017style,hu2017toward,N18-1012,P18-1080,jin2019unsupervised,chang2020dart,chang2021does,chang2020dart,chang2021neural,chang2021jointly,su2020moviechats}. 
To this end, we aim to bridge this gap that makes the following contributions: 

\begin{itemize}
    \item We showed that having ancient Chinese text of all eras in a single corpus is not ideal as they are difficult to model jointly as a single distribution, and that the additional \emph{chronological context} helps to improve translation of ancient Chinese to modern Chinese sentences.
    
    \item For future research in this direction, we release our code and parallel data consisting of annotated \emph{chronological identifiers} which allow to infer the approximate era of the written text in the practical settings.
\end{itemize}

\begin{table*}[h!]
    \centering
    \small
    \begin{tabularx}{\textwidth}{lXc}
        & Text & Chronological Period \\
        \hline
        Source (Ancient Chinese) & 孟子曰：道在尔而求诸远，事在易而求之难。 & \\
        Reference & 孟子说：道路在近旁而偏要向远处去寻求，事情本来很容易而偏要向难处下手。 (Menzie said: ``The right path is just beside but people take far away ones instead; things are easy but people handle them with difficult ways.'') & \texttt{pre-qin} \\
        System (Modern Chinese) & 孟子说：道理在于尔而求得远方，事情在于易而求得难 。 & \texttt{pre-qin} \\
        \hline 
        Source (Ancient Chinese) & 秦昭王召见，与语，大说之，拜为客卿。 & \\
        Reference & 秦昭王便召见了蔡泽，跟他谈话后，很喜欢他，授给他客卿职位。 (The King of Qin summoned Mr. Ze Cai and, after talking to him, liked him and gave him a government official position for foreigners.) & \texttt{han} \\
        System (Modern Chinese) & 秦昭王召见他，与他谈话，非常高兴，拜他为客卿。 & \texttt{han} \\
        \hline 
        Source (Ancient Chinese) & 太子曰：吾君老矣，非骊姬，寝不安，食不甘。 & \\
        Reference & 太子说：我父亲年老了，没有骊姬将睡不稳、食无味。 (The Prince said: ``My father is old. Without this girl, Li, he cannot sleep well or eat well.'') & \texttt{han} \\
        System (Modern Chinese) & 太子说：我国君已经老了，不是骊姬的姬妾，吃不甘。 & \texttt{han} \\
    \end{tabularx}
    \caption{Examples of system output consisting of the \emph{ancient Chinese source}, \emph{modern Chinese reference} and the \emph{chronological period prediction}.}
    \label{tab:output}
\end{table*}

\section{Background}

At a fine-grained view, the notion of ``ancient Chinese'' may not be considered a single language with a static word-meaning mapping.
Therefore, we direct our efforts toward three particular eras: Pre-Qin (先秦), Han (汉), and Song (宋) to verify the hypothesis that the chronology of a text directly influences the word meaning and model performance. 
In particular, \emph{Pre-Qin} and \emph{Han} are closer chronologically, so we expect their model performances to be closer than that between \emph{Pre-Qin} and \emph{Song}, as was shown in other ancient text translation~\cite{park2020ancient}.

One reason for this difference is the use of polysemous single-character words, which are highly ambiguous. 
Some words begin to lose meanings over time. 
For example, in ancient Chinese, the word 看(`kàn') has many meanings such as ``to visit'' and ``to listen'', in addition to the major modern meaning, ``to look''.

As the language evolved, vocabulary changed and lexical semantic shift took place, creating diachronic semantic gaps that may introduce subtle differences in the understanding of the text.
For instance, the earliest known meaning of ``看'' is ``to look into the distance''. 
The meaning of ``to look at something closely'' emerged during the Han period and eventually became the prominent meaning of this verb in modern Chinese.
In sum, the language change across time suggests a modeling approach that is aware of when the text was written.


\section{Task Formulation}

We assume two nonparallel datasets $A$ and $M$ of sentences in \emph{Ancient Chinese (zh-a)} and \emph{Modern Chinese (zh-m)} respectively.
A parallel dataset $P$ that contains the pairs of sentences in both variants of text is also present. 
The sizes of the three datasets are denoted as $|A|$, $|M|$ and $|P|$, respectively. 
As the nonparallel data is abundant but the parallel data is limited, size $|A|$, $|M| \gg |P|$. 
The main objective is to convert the input ancient Chinese text $a$ to its modern variant $m$. 
This task is akin to style transfer, or if the text are drastically different, machine translation. 
In this paper, we are only concerned with the direction from zh-a to zh-m.
Additionally, we include the prediction of the chronological period of the ancient text as an auxiliary task.

\section{Proposed Framework}
\label{sec:approach}

Our framework translates the given ancient Chinese text (\cref{sec:translation}) while providing additional chronological context information (\cref{sec:multitask}) (see Table~\ref{tab:output}).
We train the \emph{translation model} in a semi-supervised manner such that cheap and easy-to-obtain modern Chinese text can be utilized in the training process.
To better select from the pool of generated candidates in a time-aware way, we use the multi-label prediction model as both the \emph{reranker} and the \emph{chronology predictor}.
The predicted chronological period also provides users with crucial context for understanding the ancient text. 

\subsection{Semi-Supervised Translation Model}
\label{sec:translation}
Our sequence-to-sequence model is based on the Transformer~\cite{vaswani2017attention} encoder-decoder architecture.
Given an input, the encoder first converts it into an intermediate vector, and then the decoder takes the intermediate representation as input to generate a target output. 
In what follows, we describe the training objectives that allows the translation model to utilize augmented monolingual data.

\paragraph{Semi-Supervised Objectives.}

Inspired by the previous work on CycleGANs \citep{zhu2017unpaired} and dual learning \citep{he2016dual,chang2021jointly,chang2021neural}, our method trains the initial model in both forward and backward directions, and defines a semi-supervised optimization objective that combines direct supervision ($L_{supervised}$) and a language model loss ($L_{lm}$) over the parallel data $P$, and two monolingual corpora $A$ and $M$:
\vspace{-0.12cm}
\[L = L_{supervised}(  P   ) + L_{lm}(A) + L_{lm}(M)\]
where $L_{supervised}(  P   )$ utilizes the aligned sentence pairs in $P$ to perform {domain alignment}, ensuring that the representation of the ancient Chinese text can be semantically aligned with its modern variant.
Moreover, \emph{the semi-supervised training allows us to augment monolingual modern Chinese} for language modeling.
Empirically, we found that this benefits the forward translation from zh-a to zh-m and proves to be a viable way for improving the system.

\subsection{Multi-Label Prediction}
\label{sec:multitask}
Further, we improve upon the translation model via the use of the \emph{chronology inference} and \emph{translation reranking} via the dual-purpose \emph{multi-label prediction model}.
Specifically, we pretrain a modern Chinese language model then fine-tune this model in a task-specific manner to {help predicting the chronological period} and using it to also rank the translation model's predictions.

\paragraph{Chronology Inference.}
To do so, we first pretrain a large-scale language model on the monolingual modern Chinese corpus following objectives in~\citet{GPT2} for GPT-2.
This enables the model to be familiarized with the language semantics where some of which are transferrable to the ancient text. 
Next, we continue to train the GPT-2 model to perform conditional task-specific generation by maximizing the joint probability $p_{\mbox{\scriptsize GPT-2}}(a,m,c)$, where $a$ is the ancient Chinese text, $m$ is the modern Chinese text, and $c$ represents the contextual information as the chronological period of the ancient text.
Specifically, for each sentence pair, the ancient Chinese tokens $w^a_i$, the modern Chinese tokens $w^m_j$, and the chronological period are concatenated into ``[zh\_a] $w^a_1 \cdots w^a_{|a|}$ [zh\_m] $w^m_1 \cdots w^m_{|m|}$ [chron] $c$'', and the model is trained to maximize the probability of this sequence.

\paragraph{Quality Estimation for Reranking.}

At inference time, we append each of the chronology labels to the translation outputs, then allow the multi-label prediction model to predict their qualities. 
Specifically, the fine-tuned LM computes the negative log loss on each of the triplets $(a,m',c')$ from the upstream \emph{translation model} by appending \emph{exhaustively} all possible \emph{chronology labels} $c'$ to the end of the generated sequence $m'$ following the same format as above and selecting the best.



\section{Dataset Construction}
\label{sec:data}
We obtain parallel ancient-modern Chinese sentence pairs, and nonparallel ancient (zh-a) and modern Chinese (zh-m) sentences from two sources \cite{liu2019ancient,shang-etal-2019-semi}.
Table~\ref{tab:data_stat} summarizes the data we used for the experiments.

\paragraph{Chronology Annotation.}
In this paper, we focus on translating ancient prose.
There are a total of 28,807 ancient Chinese prose sentences.
We annotate each of these sentences with the Chinese historical period (dynasty) in which it was written.
Specifically, we consider three chronology labels: \texttt{pre-qin} (先秦), \texttt{han} (汉), and \texttt{song} (宋).
The annotation is based on the source of the sentences, i.e., which ancient book the sentences are taken from.
The total number of annotated sentences for each period is 1,244, 20,460, and 7,103 respectively.
This annotation scheme can be adopted for a larger set of periods when ancient text of a wider time span is available.


\paragraph{Parallel Data.}
For the sentences with chronology annotation, we randomly assign 10\% sentences to the development set and test set respectively.
The remaining sentences are used as training data.
We further supplement the parallel training data with 4,760 sentences from ancient Chinese poems, each also with a modern Chinese translation.
The final training, development and test set statistics and be found in Table~\ref{tab:data_stat}.
\paragraph{Nonparallel Data.}
We extend the source-side data by including 269,409 more ancient poem sentences without translation.
For extending target-side data, we add 77,687 sentences from modern lyrics, following \citet{shang-etal-2019-semi}. The details of nonparallel data are also shown in Table~\ref{tab:data_stat}.

\begin{table}[t!]
    \centering
    \resizebox{0.8\columnwidth}{!}{
    \begin{tabular}{llrc}
        & & \# sentences & \# characters \\
        \hline 
        Nonparallel & zh-a & 269,409 & 4M \\
        & zh-m & 77,687 & 826K \\
        \hline
        Parallel & Train & 27,807 & (524K, 797K) \\
        & Dev & 2,880 & (59K, 88K) \\
        & Test & 2,880 & (60K, 90K) \\
    \end{tabular}}
    \caption{Statistics of the dataset. For each part of the dataset, the number of sentences and the (source, target) number of characters are shown.}
    \label{tab:data_stat}
\end{table}

\section{Experimental Settings}
\label{sec:exp}

\begin{table}[t!]
    \centering
    \resizebox{\columnwidth}{!}{
    \begin{tabular}{lc|ccc}
        & \multicolumn{4}{c}{BLEU} \\
        Training Objectives & All & \texttt{pre-qin} & \texttt{han} & \texttt{song} \\
        \hline
        $L_{supervised}$ \cite{liu2019ancient} & 19.59 & 14.41 & \textbf{20.02} & 19.13 \\
        $L_{supervised} + L_{lm}(M)$ & 23.05 & 15.97 & \textbf{23.32} & 23.17 \\
        $L_{supervised} + L_{lm}(M) + L_{lm}(A)$ & 23.15 & 14.15 & 23.34 & \textbf{23.72} \\
        ~~+ share decoder embeddings & 24.38 & 15.70 & 24.52 & \textbf{24.99} \\
        ~~+ time-aware reranking & \textbf{24.51} & 15.50 & 24.62 & \textbf{25.24} \\ 
    \end{tabular}}
    \caption{Ancient to modern Chinese translation performance. BLEU scores are calculated with 1 to 4 character n-grams.}
    \label{tab:mt_perf}
\end{table}

We tokenized both ancient and modern Chinese text by splitting characters.
The vocabulary sizes are 4,824 and 4,600 respectively.
We built our model upon the Fairseq toolkit\footnote{\url{https://github.com/pytorch/fairseq}}.
The architecture is Transformer with about 54M parameters, which largely follows the configuration of \citet{liu2019ancient}.
Translations were generated with beam size 5, and we consider top 5 candidates for reranking.
For the \emph{multi-label prediction model}, we adapted existing code\footnote{\url{https://github.com/Morizeyao/GPT2-Chinese}} to build a GPT-2 Language Model reranker with approximately 82M parameters.
First, we pre-trained the model with 1.2 GB of Chinese Wikipedia text.
Then, we fine-tuned the pre-trained model with the chronologically-annotated training data.
For each ancient-modern sentence pair with chronology information, we form a text-period query string with the scheme described in~\cref{sec:multitask}.
We select the final model according to perplexity computed on the development set.

\section{Main Results}
\label{sec:main}

Overall, we observe from Table~\ref{tab:mt_perf} and~\ref{tab:chron_perf} that the use of the multi-label prediction model not only allows for better context than pure translation, but also helps to boost the general performance on the translation tasks. 
Moreover, \emph{translations of ancient text chronologically closer to modern Chinese (\texttt{han} and \texttt{song}) tend to yield better performances, as the semantic gaps are generally smaller}.
We also demonstrate that the semi-supervised training which avail of the additional nonparallel text helps to improve the translation model even further.
Specifically, zh-m nonparallel data enhances the decoder's ability to generate modern Chinese, while zh-a nonparallel data may help the encoder to maintain crucial semantic information.
We achieved a BLEU score of $23.15$ in this setting.
As the source and target side vocabularies have a large overlap, we experimented with sharing decoder embeddings and got +$1.23$ BLEU improvement, which may also serve as an evidence that there are still ancient components in modern Chinese.
Finally, reranking further boosted the BLEU score to $24.51$.

\begin{table}[t]
    \centering
    \small
    \resizebox{0.75\columnwidth}{!}{
    \begin{tabular}{r|ccc}
        Period (\# test) & Precision & Recall & F1 \\
        \hline 
        \texttt{pre-qin} (117) & 0.05 & 0.53 & 0.09 \\
        \texttt{han} (2043) & 0.85 & 0.57 & 0.68 \\
        \texttt{song} (720) & 0.85 & 0.27 & 0.41 \\
        \hline
        Accuracy & & & 0.49 \\
        Macro avg. & 0.58 & 0.45 & 0.39 \\
        Weighted avg. & 0.82 & 0.49 & 0.59 \\
    \end{tabular}}
    \caption{Performance of Chronology Inference}
    \label{tab:chron_perf}
\end{table}
\begin{figure}[t]
    \centering
    \resizebox{\columnwidth}{!}{
    \includegraphics[width=\columnwidth]{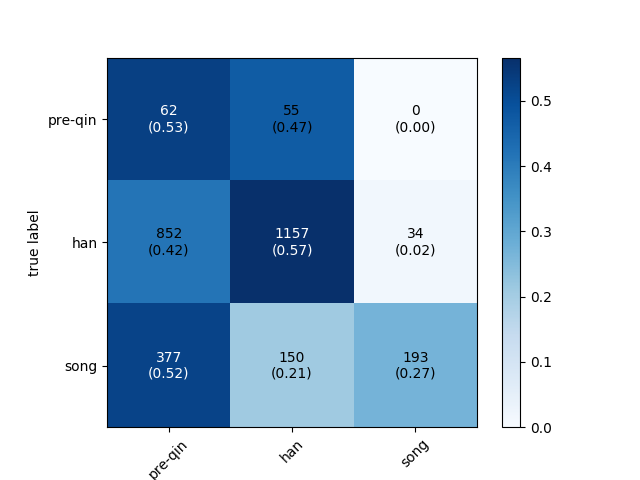}
    }
    \caption{Confusion matrix for Chronology Inference}
    \label{fig:chron_conf_mat}
\end{figure}

\paragraph{Error Analysis.}
\label{sec:error_analysis}

We perform \emph{human evaluation} on $100$ randomly sampled output instances and observe them to be high in \emph{adequacy} and \emph{fluency}, $4.06$ and $3.68$ respectively, on a scale of $0$-$5$.
This was done by averaging the fluency and adequacy ratings of three domain experts.
Further, we also observe that the chronology of text impacts the model performance as in Table~\ref{tab:mt_perf}. 
Leveraging zh-m nonparallel data is most helpful for translating text from the \texttt{song} period, which is much closer to modern Chinese compared to the text from the other two periods.
Further, from Figure~\ref{fig:chron_conf_mat} we observe that the chronology inference depends very much on the \emph{data scarcity} and the \emph{closeness} of chronological periods. 
On the Chinese historical timeline, \texttt{han} is very close to \texttt{pre-qin}, but \texttt{han} and \texttt{song} are more separated.
Another source of difficulty is that ancient Chinese writings tend to quote a considerable amount of text written in previous time periods.
For example, a history book written in the \texttt{song} period may inherit narratives written in \texttt{pre-qin} and \texttt{han} for the history before \texttt{han}.
As a result, it is challenging to perform chronology inference based solely on the linguistic properties of individual sentences.
Nevertheless, chronological inference can still provide useful signals for the translation model to better capture semantic differences across time.

\section{Conclusion}

In this paper, we present a framework that translates ancient Chinese texts into its modern correspondence in low resource scenarios with very little parallel data and a larger set of nonparallel sentences without ancient-modern alignment information.
We display the importance and usefulness of chronology inference as an auxiliary task that hints at potential diachronic semantic gaps.
We hope to extend this research to further model additional contextual information about each era.

\section*{Acknowledgements}
This research was funded in part by the German Research Foundation (DFG) as part of SFB 248 ``Foundations of Perspicuous Software Systems''. We sincerely thank the anonymous reviewers for their insightful comments that helped us to improve this paper.

\bibliography{anthology,acl2021}

\bibliographystyle{acl_natbib}

\appendix
\counterwithin{table}{section}
\clearpage
\onecolumn


\end{CJK}

\end{document}